\documentclass[11pt]{article}

\usepackage[final]{acl}

\usepackage{times}
\usepackage{latexsym}
\usepackage[T1]{fontenc}
\usepackage[utf8]{inputenc}
\usepackage{microtype}
\usepackage{inconsolata}
\usepackage{graphicx}
\usepackage{array}
\usepackage{xcolor}
\usepackage{tikz}
\usepackage{pgfplots}
\usepackage{booktabs}
\usepackage{listings}
\usetikzlibrary{arrows.meta}
\pgfplotsset{compat=1.18}

\newcommand{\model}{Nemotron 3.5 Content Safety Moderator}

\newcommand{\modeltable}{Nemotron\allowbreak{} 3.5 CS}

\newcommand{\reasonmark}{\textsuperscript{$\dagger$}}

\newcolumntype{L}[1]{>{\raggedright\arraybackslash}p{#1}}
\newcolumntype{C}[1]{>{\centering\arraybackslash}p{#1}}
\lstdefinestyle{promptlisting}{
    basicstyle=\ttfamily\scriptsize,
    breaklines=true,
    columns=fullflexible,
    keepspaces=true,
    showstringspaces=false
}

\title{\model{}: A Compact Multimodal, Multilingual, and Reasoning Enabled Content Safety Moderator}
\author{Varun Singh, Anuj Doshi, Makesh Narsimhan Sreedhar, Shaona Ghosh and Katherine Luna \\
        NVIDIA \\
        Santa Clara, CA \\
        \texttt{\{vasingh, andoshi, makeshn, shaonag, kluna\}@nvidia.com}
}

\begin{document}
\maketitle
\begin{abstract}
Safety moderation for deployed AI applications is moving beyond text-only prompts: systems increasingly need to judge images, documents, screenshots, and generated responses under policies that vary across domains. Existing guardrails usually cover only part of this setting, making it difficult to combine broad coverage, custom policy control, and low compute cost. We present \model{}, also referred to as \modeltable{} in this paper for brevity, a compact 4B vision-language safety moderator that jointly classifies user prompts, images, and assistant responses across 12 languages. \modeltable{} returns safety labels for latency-sensitive moderation and can additionally produce concise reasoning traces that apply supplied custom policies and identify violated categories when reasoning is requested. We also release a multimodal and multilingual safety dataset for guard training, spanning human-labeled real-image moderation, benign vision-language and document tasks, synthetic rare-risk and jailbreak cases, and custom-policy examples. Across evaluations spanning multimodal safety, text moderation, multilingual robustness, custom-policy following, benign false positives, and latency, \modeltable{} demonstrates a practical coverage tradeoff: it adds image-conditioned and policy-conditioned moderation while remaining broadly competitive with specialized guard models. These results suggest that compact vision-language moderators can serve as deployable front-line safety components, with reasoning used selectively for audit and policy review.
\end{abstract}

\section{Introduction}

Safety moderation has become a standard component of deployed language-model systems. Recent guard models and datasets have improved prompt harmfulness detection, response harmfulness detection, and refusal analysis for text interactions \citep{han-etal-2024-wildguard,ghosh-etal-2025-aegis2}. This progress, however, does not fully cover applications whose safety decisions depend on visual context, non-English inputs, generated responses, or policies that differ across domains.

Multimodal safety work shows that visual information can change whether an instruction is safe and that image-conditioned attacks can bypass text-only safeguards \citep{liu-etal-2023-mmsafetybench,zong-etal-2024-vlguard}. Multilingual safety work similarly finds that English-centered moderation leaves substantial gaps for global deployments \citep{kumar-etal-2025-polyguard}. Recent safety evaluation frameworks further show that application-specific policies and over-refusal patterns are difficult to capture with fixed taxonomies alone \citep{jindal-etal-2025-sage}. These findings point to a practical gap: moderation systems increasingly need to inspect text, images, and responses together, apply supplied custom policies, and still run fast enough for latency-sensitive use.

\begin{figure}[t]
\centering
\includegraphics[width=\columnwidth]{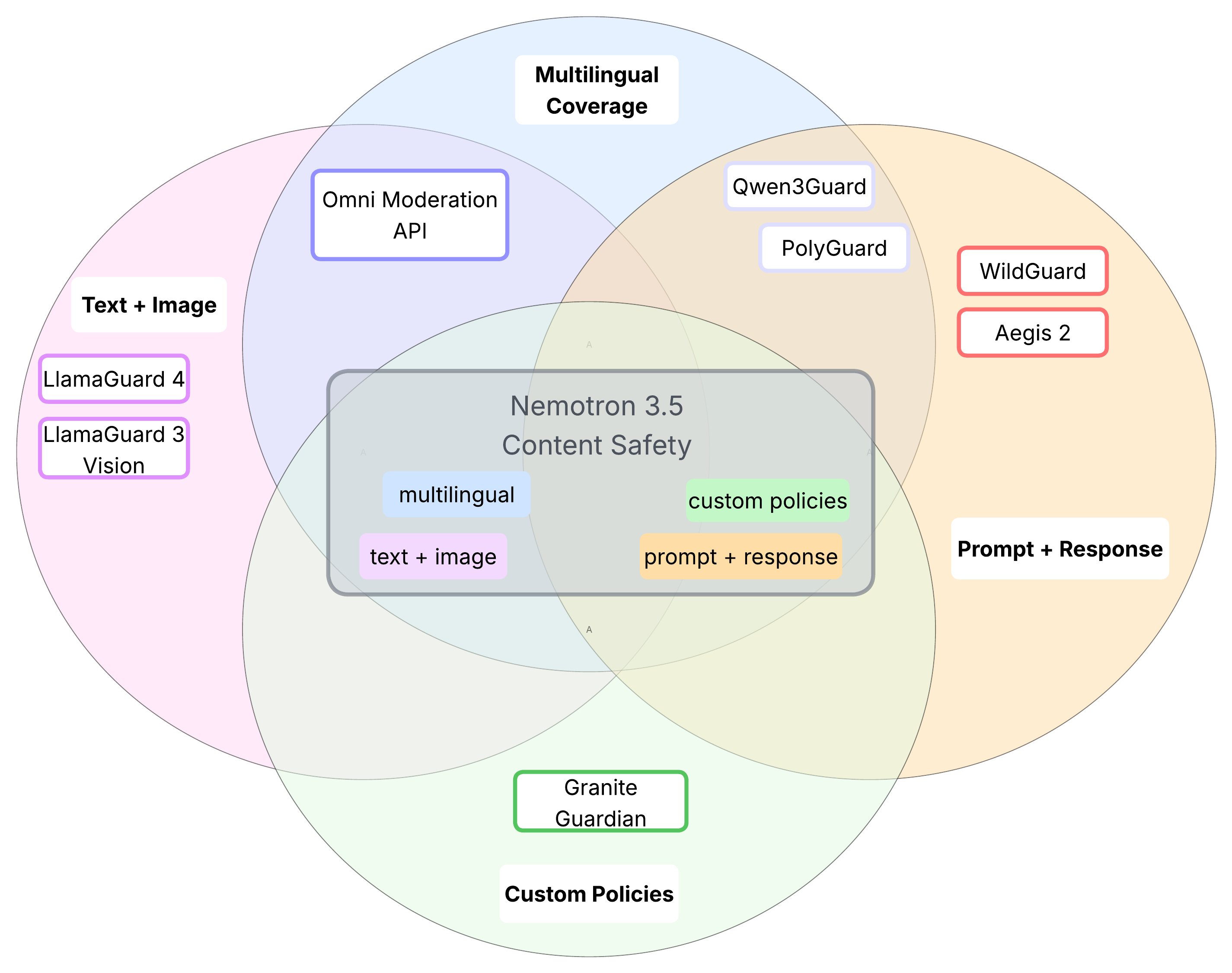}
\caption{Landscape of guard models. Colored regions denote supported capabilities, and boxes denote example systems. \modeltable{} is designed to combine image-conditioned moderation, prompt and response classification, multilingual coverage, and custom-policy support in one compact model.}
\label{fig:moderators-landscape}
\end{figure}

Figure~\ref{fig:moderators-landscape} illustrates this gap: existing guards tend to specialize along one or two dimensions, while deployed applications often need the intersection of multimodal inputs, response-side checks, multilingual coverage, and policy-specific behavior. We introduce \model{}, a compact 4B vision-language content-safety moderator designed for this setting. The model classifies user prompts, images, and assistant responses when they are present in a single context; supports 12 explicitly trained languages; and accepts custom policies at inference time. For latency-sensitive checks, it can return safety labels directly. When reasoning is requested, it returns concise reasoning traces with violated categories for audit or policy review. Rather than treating moderation as a fixed standalone classifier, we design the interface, data mixture, and evaluation around the constraints that make safety filters usable in practice: multimodal coverage, multilingual reliability, low false positives on benign inputs, custom-policy flexibility, and low inference latency.

Our contributions are:
\begin{enumerate}
    \item We train a compact 4B multimodal safety moderator for joint prompt, image, and response classification across 12 explicitly trained languages, with reasoning capability for supplied custom policies.
    \item We curate and release a multimodal and multilingual safety dataset spanning human-labeled real-image examples, benign vision-language examples, synthetic rare-risk and jailbreak examples, and policy-following data, while also describing the training recipe built from it.
    \item We present our methodology for generating synthetic data that can be used for generating training and evaluation datasets.
\end{enumerate}

\section{Related Work}
\paragraph{Text-only safety models.}
Text-only guard models established LLM-based prompt and response moderation, from the Llama Guard family \citep{inan-etal-2023-llamaguard,metallamateam-2024-llamaguard2} to broader safety classifiers such as WildGuard, AEGIS2, and ShieldGemma \citep{han-etal-2024-wildguard,ghosh-etal-2025-aegis2,zeng-etal-2024-shieldgemma}. These systems provide strong text baselines, but do not jointly address visual context, multilingual gaps, or custom-policy flexibility.

\paragraph{Multilingual safety.}
Multilingual safety benchmarks show that English-centered alignment often fails under translated, low-resource, or culturally specific harmful prompts \citep{wang-etal-2023-xsafety,deng-etal-2024-multijail,degibert-etal-2024-rtplx,ahmadian-etal-2024-aya}. PolyGuard and Qwen3Guard further demonstrate the value of multilingual guard training \citep{kumar-etal-2025-polyguard,qwenteam-2025-qwen3guard}, but remain primarily text-only and do not address image-conditioned moderation.

\paragraph{Multimodal safety.}
Multimodal safety work shows that harmful intent can depend on image-text interaction or be hidden in visual prompts \citep{liu-etal-2023-mmsafetybench,gong-etal-2023-figstep,zong-etal-2024-vlguard}. Recent multimodal guards, including Llama Guard 3 Vision and Llama Guard 4, extend safety classification to image-conditioned inputs and responses \citep{chi-etal-2024-llamaguard3vision,metallamateam-2025-llamaguard4}, while MSTS highlights the remaining multilingual multimodal safety gap \citep{rottger-etal-2025-msts}.

\paragraph{Policy-conditioned and reasoning-based safety.}
Policy-conditioned safety instead treats moderation as instruction following over supplied rules \citep{jindal-etal-2025-sage,gupta-etal-2024-granite,qwenteam-2025-qwen3guard}. These designs improve deployment flexibility, but existing systems do not combine policy conditioning with both multimodal inputs and multilingual coverage.

\section{Methodology}
\modeltable{} is designed around four requirements. First, both the model weights and training data must be openly released under permissive licenses, enabling community reuse and reproducibility. Second, the model must have a small enough parameter count to be deployable in compute-constrained settings, including edge inference. Third, it must support multimodal inputs so it can serve as a guard for the growing class of VLM-based applications. Fourth, it must provide reliable coverage across multiple languages to be usable in global deployments.

We selected the Gemma 3-4B model \citep{gemmateam-2025-gemma3} as our base as it fulfills all the requirements. 

\paragraph{Input and Output Interface}

\modeltable{} accepts a structured context assembled from up to three components: (1)~a \textit{user input}, consisting of text with an optional accompanying image; (2)~an \textit{assistant response}, enabling response-side moderation; and (3)~a \textit{custom policy}, a free-form text specification of domain-specific safety requirements or permitted content. We provide a custom chat template that assembles these components into a single sequence; Appendix~\ref{app:chat-template} gives the template and default prompt.

\paragraph{Safety Taxonomy}
\label{sec:taxonomy}

Our taxonomy is adapted from the AEGIS 2.0 safety taxonomy \citep{ghosh-etal-2025-aegis2}, which defines thirteen core unsafe categories alongside fine-grained extended categories covering illegal activity, fraud and deception, manipulation, malware, and unauthorized advice. We add one additional fine-grained category, economic harm, to cover financial fraud, predatory lending, market manipulation, and related harms not explicitly represented in the base taxonomy.

\paragraph{Fine-tuning Procedure}
\label{sec:finetuning}

We fine-tune the Gemma 3-4B base model using supervised fine-tuning (SFT) implemented with LlamaFactory \citep{zheng-etal-2024-llamafactory}, an open-source unified training framework. Each training example is formatted as an input/output pair where the input contains the full classification context: the user prompt, optional image, optional assistant response, the think mode token (\texttt{/think} or \texttt{/no\_think}), and the category-presence token (\texttt{/categories} or \texttt{/no\_categories}). The output contains the safety label of the input and output and the violated category list if either the input or output is unsafe; for reasoning examples a chain-of-thought reasoning trace is prepended to the output. All training is conducted on NVIDIA H100 GPUs. We report SFT hyperparameters in Appendix~\ref{app:sft-hyperparams}.

\paragraph{Inference}
\label{sec:inference}
\modeltable{} supports two inference modes. In \textit{direct classification mode}, the model emits a binary \textsc{safe}/\textsc{unsafe} label and, for unsafe inputs, an optional list of violated categories drawn from our taxonomy. This mode minimizes time-to-first-token and is designed as a front-line filter in latency-sensitive pipelines. In \textit{reasoning mode}, the model first produces a concise chain-of-thought trace that applies any supplied custom policy and cites specific violated categories, before emitting the final verdict. This mode supports policy review and audit workflows where the rationale behind a moderation decision must be surfaced to an operator or downstream system. The two inference modes can be toggled at inference time by appending the mode token (\texttt{/think} or \texttt{/no\_think}) to the user input. In each of the inference modes, the model can also be toggled to emit a list of violated categories alongside the safety verdict using the category-presence token (\texttt{/categories} or \texttt{/no\_categories}).

\section{Data and Training}
\label{sec:data}
Curating safety training data is one of the most demanding aspects of building a safety moderator because several properties make it harder than general-purpose supervised learning:
\begin{itemize}
\setlength{\itemsep}{0pt}
\setlength{\parsep}{0pt}
\setlength{\topsep}{2pt}
\setlength{\partopsep}{0pt}
    \item \textbf{Domain specificity:} what counts as harmful varies significantly across deployment contexts, so a dataset for one application may be poorly calibrated for another.
    \item \textbf{Taxonomy fragmentation:} there is no community-wide standard for harm categories; each dataset and guard model defines its own label space, making cross-dataset aggregation non-trivial.
    \item \textbf{Content sensitivity:} safety data contains harmful, profane, or otherwise objectionable material, complicating open distribution and imposing welfare requirements on annotators.
    \item \textbf{Legal constraints on multimodal data:} images depicting real people, copyrighted works, or regulated content may carry licensing restrictions that limit redistribution.
    \item \textbf{Annotation cost:} labeling requires specialized judgment, often backed by trained reviewers, making it substantially more expensive per example than general NLP annotation.
\end{itemize}

To address these challenges, our training mixture combines human annotation, reuse of existing public datasets, and synthetic data generation (SDG). Human-labeled multimodal examples provide grounded coverage of real-world visual safety scenarios. Public text-only safety datasets are incorporated, taking advantage of well-established datasets. Synthetically generated examples fill coverage gaps for rare harm categories and adversarial patterns. Finally, we include topic-following examples following \citet{sreedhar-etal-2025-safety}, who show that topic-following is a generalized form of content moderation; these examples teach the model to apply operator-supplied custom policies at inference time.

\paragraph{Human Annotation}
Our human annotation pipeline consists of two high level steps: \textit{a.}~sourcing of data and \textit{b.}~labeling of data. Our images are sourced from internal and external sources subject to varying licensing restrictions. We utilized in-house annotators to label the data. For more details on our human annotation project setups, see Appendix~\ref{app:annotation-setups}.

\paragraph{Training Data Mixture and Recipe}

Table~\ref{tab:data-mixture} summarizes the six source types in the final training mixture. One notable property of the human-annotated multimodal subset is that \textbf{99\% of training images are real photographs or text on a plain background}, not synthetic generations. This directly addresses a known weakness of existing multimodal safety datasets such as VLGuard \citep{zong-etal-2024-vlguard} and MM-SafetyBench \citep{liu-etal-2023-mmsafetybench}, which rely heavily on synthetic images that are often blurry, have low resolution, and lack the cultural texture of production content.

\begin{table}[h]
\centering
\small
\begin{tabular}{p{2.4cm} p{4.6cm}}
\hline
\textbf{Source} & \textbf{Description} \\
\hline
Human-annotated multimodal (multiple sources including Wikimedia) & In-house collected English data translated into 12 languages along with chain-of-thought outputs distilled from Qwen 397B, shortened with Qwen 80B \\
Nemotron Safety Guard Dataset v3 & Multilingual text safety data; proportional coverage across categories and safe/unsafe splits in 12 languages \\
Nemotron VLM Dataset v2 & Safe multimodal data (documents, charts, diagrams); reduces over-refusal on benign professional content \\
CantTalk\-About\-This & Topic-following data; policy/verdict pairs across healthcare, finance, education, and similar domains \\
Synthetic generation & Synthetically generated data for jailbreak variants, rare policy violations, and refusals (Section~\ref{sec:sdg}) \\
\hline
\end{tabular}
\caption{Released dataset composition. Sources are listed in approximate descending order of volume contribution.}
\label{tab:data-mixture}
\end{table}

\section{Synthetic Data Generation (SDG)}
\label{sec:sdg}
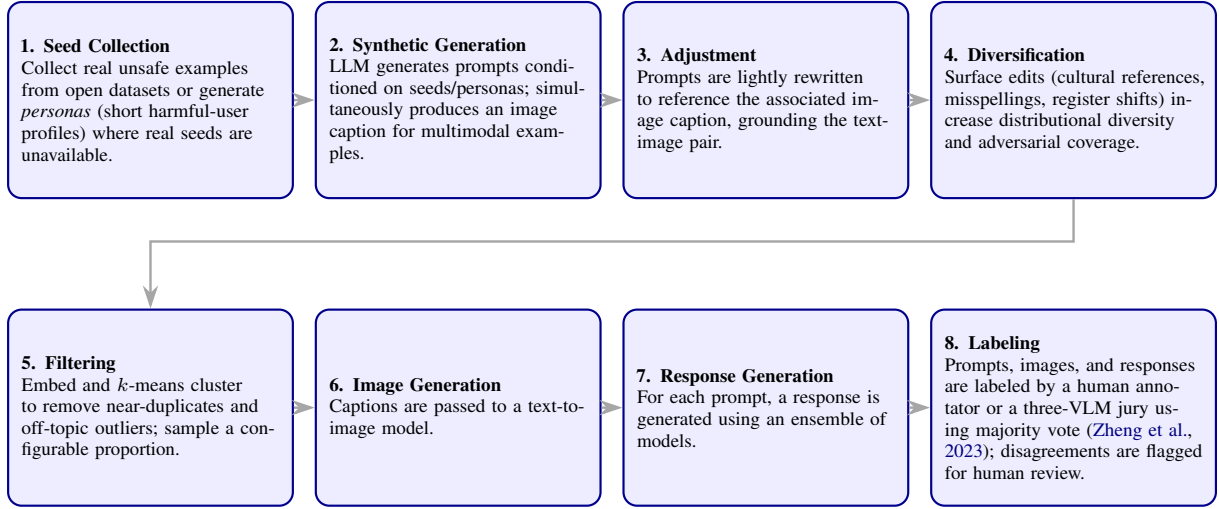
\begin{figure*}[t]
\centering
\scriptsize
\resizebox{\textwidth}{!}{%
\begin{tikzpicture}[
    stagebase/.style={
        rounded corners=5pt,
        line width=0.8pt,
        text width=3.35cm,
        minimum height=2.55cm,
        inner sep=5pt,
        align=left
    },
    stage/.style={stagebase, draw=blue!55!black, fill=blue!7},
    arrow/.style={-{Stealth[length=3mm,width=2mm]}, line width=0.9pt, draw=gray!70}
]
\node[stage] (s1) at (0,4.0) {\textbf{1. Seed Collection}\\Collect real unsafe examples from open datasets or generate \textit{personas} (short harmful-user profiles) where real seeds are unavailable.};
\node[stage] (s2) at (4.0,4.0) {\textbf{2. Synthetic Generation}\\LLM generates prompts conditioned on seeds/personas; simultaneously produces an image caption for multimodal examples.};
\node[stage] (s3) at (8.0,4.0) {\textbf{3. Adjustment}\\Prompts are lightly rewritten to reference the associated image caption, grounding the text-image pair.};
\node[stage] (s4) at (12.0,4.0) {\textbf{4. Diversification}\\Surface edits (cultural references, misspellings, register shifts) increase distributional diversity and adversarial coverage.};

\node[stage] (s5) at (0,0) {\textbf{5. Filtering}\\Embed and $k$-means cluster to remove near-duplicates and off-topic outliers; sample a configurable proportion.};
\node[stage] (s6) at (4.0,0) {\textbf{6. Image Generation}\\Captions are passed to a text-to-image model.};
\node[stage] (s7) at (8.0,0) {\textbf{7. Response Generation}\\For each prompt, a response is generated using an ensemble of models.};
\node[stage] (s8) at (12.0,0) {\textbf{8. Labeling}\\Prompts, images, and responses are labeled by a human annotator or a three-VLM jury using majority vote \citep{zheng-etal-2023-judging}; disagreements are flagged for human review.};

\draw[arrow] (s1) -- (s2);
\draw[arrow] (s2) -- (s3);
\draw[arrow] (s3) -- (s4);
\draw[arrow] (s4.south) -- ++(0,-0.55) -| (s5.north);
\draw[arrow] (s5) -- (s6);
\draw[arrow] (s6) -- (s7);
\draw[arrow] (s7) -- (s8);
\end{tikzpicture}}
\caption{Eight-stage SDG pipeline for producing multimodal safety training data. Each box denotes one stage, and arrows indicate the flow from seed collection through labeling.}
\label{fig:sdg-pipeline}
\end{figure*}

SDG supplements training data for harm categories that are scarce, legally restricted, or unsafe for annotators to produce directly. We apply it primarily to generate synthetic refusals and jailbreak patterns. Our eight-stage pipeline is summarized in Figure~\ref{fig:sdg-pipeline}.

\section{Evaluation}

We evaluate \modeltable{} along six capabilities: multimodal harmful-content detection, text safety, multilingual robustness, custom-policy following, benign-input false positives, and latency. Each capability answers a distinct evaluation question, so we report the metric aligned with the benchmark purpose and provide fuller metric, language, and category breakdowns in Appendix~\ref{sec:additional-results}.

\paragraph{Models and baselines.}
We compare against baseline families with different interface coverage, including multimodal guards, text-only safety classifiers, multilingual guard models, and custom-policy systems. Since image inputs, response-side classification, custom policies, and latency measurement are not supported by every system, we use the strongest applicable comparison set for each capability; unsupported settings are marked in Appendix~\ref{sec:additional-results}.

\paragraph{Metrics.}
Our primary classification metric is harmful-F1 over the unsafe class. We additionally report harmful recall, benign false-positive rate, and latency when those metrics better match the benchmark purpose; benign vision-language sets are treated as safe inputs and scored by false-positive rate. Latency is measured with time to first token and end-to-end response time.

\paragraph{Multimodal safety.}
VLGuard and MM-SafetyBench test unsafe requests whose interpretation depends on both image and text. We report prompt-side accuracy and harmful-F1 for VLGuard, harmful-F1 for MM-SafetyBench, and category-level results in Appendix~\ref{sec:additional-results}.

\paragraph{Text safety benchmarks.}
We evaluate prompt and response classification on Aegis 2.0, XSTest~\citep{rottger-etal-2023-xstest}, and WildGuard to check whether the multimodal model remains competitive on established text moderation tasks.

\paragraph{Multilingual safety.}
We evaluate multilingual safety on PolyGuard, RTP-LX, MultiJail, XSafety, Aya Red Teaming, Multilingual Aegis, and LinguaSafe~\citep{ning-etal-2025-linguasafe}.

\paragraph{Custom-policy following and reasoning.}
We evaluate custom-policy following with DynaGuardrail~\citep{hoover-etal-2025-dynaguard} and CoSA~\citep{zhang-etal-2025-cosa}, which require applying supplied policy descriptions rather than a fixed taxonomy. We report F1 by DynaGuardrail policy domain and CoSA scenario, including both direct and reasoning-on variants for \modeltable{}.

\paragraph{Benign multimodal false positives.}
We measure overblocking on MMMU~\citep{yue-etal-2023-mmmu}, DocVQA~\citep{mathew-etal-2021-docvqa}, and AI2D~\citep{kembhavi-etal-2016-ai2d} by treating examples as safe inputs and reporting false-positive rate. These benchmarks cover screenshots, forms, documents, charts, and educational diagrams that a useful moderator should not block.

\paragraph{Latency.}
We measure latency on RTVLM~\citep{li-etal-2024-rtvlm} image-text inputs and DynaGuardrail custom-policy inputs, separating direct classification from reasoning-enabled outputs.

\section{Results and Analysis}

Table~\ref{tab:capability-dashboard} summarizes the main results, with detailed model-by-benchmark matrices in Appendix~\ref{sec:additional-results}.

\begin{table*}[t]
    \centering
    \small
    \renewcommand{\arraystretch}{1.08}
    {\setlength{\tabcolsep}{2.2pt}
    \begin{tabular*}{\textwidth}{@{\extracolsep{\fill}}L{2.0cm}L{2.55cm}L{1.2cm}L{1.65cm}C{1.4cm}C{1.2cm}C{1.35cm}C{1.85cm}@{}}
    \hline
    Capability & Evaluation & Slice & Metric & \modeltable{} & \begin{tabular}[c]{@{}c@{}}Best\\baseline\end{tabular} & $\Delta$ & Baseline \\
    \hline
    Multimodal safety & VLGuard, MM-SafetyBench & Prompt & Avg. unsafe F1 $\uparrow$ & \textbf{0.81} & 0.68 & +0.13 & Gemma 3-4B \\
    Text safety & Aegis 2.0, XSTest, WildGuard & Prompt & Avg. unsafe F1 $\uparrow$ & 0.85 & \textbf{0.89} & -0.04 & Qwen3G \\
    Text safety & Aegis 2.0, XSTest, WildGuard & Response & Avg. unsafe F1 $\uparrow$ & 0.83 & \textbf{0.86} & -0.03 & Qwen3G \\
    Multilingual safety & 7 multilingual benchmarks & Prompt & Avg. unsafe F1 $\uparrow$ & \textbf{0.854} & 0.853 & +0.001 & Qwen3G \\
    Multilingual safety & PolyGuard, Multilingual Aegis & Response & Avg. unsafe F1 $\uparrow$ & \textbf{0.847} & 0.817 & +0.030 & Qwen3G \\
    Custom policy & DynaGuardrail, CoSA & Prompt & Avg. unsafe F1 $\uparrow$ & \textbf{0.85} \newline \textbf{0.85}\reasonmark{} & 0.83 & +0.02 & GG3.3 \\
    Benign multimodal & MMMU, DocVQA, AI2D & Input & FPR $\downarrow$ & 0.030 & \textbf{0.026} & +0.004 & LG4 \\
    Latency & RTVLM & Image-text & TTFT/E2E $\downarrow$ & \textbf{60}/\textbf{76} & 99/118 & \mbox{-39/-41} & LG4 \\
    Latency & DynaGuardrail avg. & Text policy & TTFT/E2E $\downarrow$ & \textbf{17}/\textbf{33} & 21/44 & \mbox{-4/-11} & LG4 \\
    \hline
    \end{tabular*}}
    \caption{Capability scorecard. Each row compares \modeltable{} with the strongest applicable baseline for that slice. $\uparrow$ means higher is better; $\downarrow$ means lower is better. $\Delta$ is \modeltable{} minus the best baseline, so negative is favorable for FPR and latency. Latency values are milliseconds. Qwen3G = Qwen3Guard, GG3.3 = Granite Guardian 3.3, LG4 = Llama Guard 4, and \reasonmark{} = reasoning on.}
    \label{tab:capability-dashboard}
\end{table*}

\paragraph{Safety coverage.}
\modeltable{} is strongest where moderation requires image context, improving over multimodal-capable baselines while staying competitive on text safety. Qwen3Guard remains stronger on the prompt and response splits of established text-only safety benchmarks. The result is breadth: a compact multimodal moderator adds visual safety coverage while retaining strong text behavior.

\paragraph{Multilingual performance.}
The multilingual results show broad coverage across the 12 trained languages: prompt-side performance is near parity with Qwen3Guard, while response-capable multilingual benchmarks favor \modeltable{}. The model shows consistent performance across languages even for out-of-training-set languages (e.g., Vietnamese and Russian on LinguaSafe). Appendix~\ref{sec:additional-results} reports per-language tables.

\paragraph{Custom policy and reasoning.}
Appendix Tables~\ref{tab:dynaguardrail} and~\ref{tab:cosa} show that \modeltable{} can apply supplied custom policies across domain-specific settings. Outputs with reasoning on support inspection, but do not uniformly improve F1, so we report them separately rather than as a higher-scoring default.

\paragraph{Benign false positives.}
Appendix Table~\ref{tab:benign-fpr} shows low false-positive rates on MMMU and AI2D, with DocVQA as the main failure slice. Document-like inputs can resemble privacy-sensitive content, making this a priority for targeted data improvements.

\paragraph{Latency.}
At max concurrency 1, \modeltable{} reaches 60/76 ms TTFT/E2E on RTVLM image-text inputs, compared with 99/118 ms for Llama Guard 4, and averages about 17/33 ms on DynaGuardrail text/custom-policy inputs. Reasoning on increases E2E latency because it emits traces; Appendix Figure~\ref{fig:latency-throughput} shows the concurrency tradeoff.

\section{Conclusion}

We presented \model{}, a compact 4B vision-language safety moderator that jointly classifies prompts, images, and assistant responses across 12 languages, supports supplied custom policies, and offers optional reasoning traces for audit. The results show that a single compact model can add image-conditioned and policy-conditioned moderation while remaining competitive with specialized text guards. We release model weights, training data, and the SDG pipeline to support further work on deployable, open-weights multimodal safety moderation.

\section{Ethical Considerations}
This work relies on safety datasets that include human judgments about harmful, and benign content. Such labels can reflect annotators' cultural backgrounds, policy interpretations, and fatigue, even when annotation guidelines and quality checks are in place. The multimodal data also has limits in visual diversity because images are drawn from a small number of approved sources, which may overrepresent particular image styles, geographies, demographics, document formats, and everyday settings. Together, these factors can lead the model to over-block some topics or communities while under-detecting harms expressed in less represented dialects, languages, or cultural settings. Downstream users should therefore audit model behavior against their own policies, user populations, and deployment contexts.

Releasing safety data also creates dual-use risk. The same examples that help researchers train and evaluate guard models may help adversaries infer category boundaries, design jailbreaks, or train models to generate or disguise harmful content. We release the data to support transparent research on moderation, but it should be handled as a sensitive artifact.

\section{Limitations}

\modeltable{} currently accepts a single image alongside text, so videos, multi-page documents, and audio inputs remain out of scope. Document-like inputs such as forms, scanned pages, and charts remain the main source of false positives, particularly on DocVQA, and would benefit from targeted data improvements. Public benchmarks with image-conditioned response-side safety labels are scarce, limiting evaluation in multimodal settings; aggregate scores can also mask language- and category-level failures, so Appendix~\ref{sec:additional-results} reports more detailed slices. Reasoning mode increases latency because it emits longer traces, making direct classification the intended path for latency-sensitive deployments. Finally, some training and evaluation data cannot be released due to licensing and privacy restrictions, limiting direct reproducibility for portions of the training mixture.

\bibliography{custom}

@inproceedings{sreedhar-etal-2025-safety,
    title = "Safety Through Reasoning: An Empirical Study of Reasoning Guardrail Models",
    author = "Sreedhar, Makesh Narsimhan  and
      Rebedea, Traian  and
      Parisien, Christopher",
    editor = "Christodoulopoulos, Christos  and
      Chakraborty, Tanmoy  and
      Rose, Carolyn  and
      Peng, Violet",
    booktitle = "Findings of the Association for Computational Linguistics: EMNLP 2025",
    month = nov,
    year = "2025",
    address = "Suzhou, China",
    publisher = "Association for Computational Linguistics",
    url = "https://aclanthology.org/2025.findings-emnlp.1193/",
    pages = "21862--21880",
    ISBN = "979-8-89176-335-7"
}

@inproceedings{zheng-etal-2024-llamafactory,
  title        = {{LlamaFactory}: Unified Efficient Fine-Tuning of 100+ Language Models},
  author       = {Zheng, Yaowei and Zhang, Richong and Zhang, Junhao and Ye, Yanhan and Luo, Zheyan and Feng, Zhangchi and Ma, Yongqiang},
  booktitle    = {Proceedings of the 62nd Annual Meeting of the Association for Computational Linguistics (Volume 3: System Demonstrations)},
  pages        = {400--410},
  year         = {2024},
  publisher    = {Association for Computational Linguistics},
  url          = {https://aclanthology.org/2024.acl-demos.38},
}

@techreport{gemmateam-2025-gemma3,
  title        = {Gemma 3 Technical Report},
  author       = {{Google DeepMind}},
  institution  = {Google DeepMind},
  note         = {arXiv preprint arXiv:2503.19786},
  year         = {2025}
}

@inproceedings{zheng-etal-2023-judging,
  title        = {Judging {LLM}-as-a-Judge with {MT}-Bench and Chatbot Arena},
  author       = {Zheng, Lianmin and Chiang, Wei-Lin and Sheng, Ying and Zhuang, Siyuan and
                  Wu, Zhanghao and Zhuang, Yonghao and Lin, Zi and Li, Zhuohan and
                  Li, Dacheng and Xing, Eric P. and Zhang, Hao and Gonzalez, Joseph E. and
                  Stoica, Ion},
  booktitle    = {Advances in Neural Information Processing Systems},
  volume       = {36},
  year         = {2023}
}

@misc{han-etal-2024-wildguard,
  title = {{WildGuard}: Open One-Stop Moderation Tools for Safety Risks, Jailbreaks, and Refusals of {LLM}s},
  author = {Han, Seungju and Rao, Kavel and Ettinger, Allyson and Jiang, Liwei and Lin, Bill Yuchen and Lambert, Nathan and Choi, Yejin and Dziri, Nouha},
  year = {2024},
  url = {https://arxiv.org/abs/2406.18495},
  note = {arXiv:2406.18495}
}

@misc{ghosh-etal-2025-aegis2,
  title = {{Aegis2.0}: A Diverse {AI} Safety Dataset and Risks Taxonomy for Alignment of {LLM} Guardrails},
  author = {Ghosh, Shaona and Varshney, Prasoon and Sreedhar, Makesh Narsimhan and Padmakumar, Aishwarya and Rebedea, Traian and Varghese, Jibin Rajan and Parisien, Christopher},
  year = {2025},
  url = {https://arxiv.org/abs/2501.09004},
  note = {arXiv:2501.09004}
}

@misc{liu-etal-2023-mmsafetybench,
  title = {{MM-SafetyBench}: A Benchmark for Safety Evaluation of Multimodal Large Language Models},
  author = {Liu, Xin and Zhu, Yichen and Gu, Jindong and Lan, Yunshi and Yang, Chao and Qiao, Yu},
  year = {2023},
  url = {https://arxiv.org/abs/2311.17600},
  note = {arXiv:2311.17600}
}

@misc{zong-etal-2024-vlguard,
  title = {Safety Fine-Tuning at (Almost) No Cost: A Baseline for Vision Large Language Models},
  author = {Zong, Yongshuo and Bohdal, Ondrej and Yu, Tingting and Yang, Yongxin and Hospedales, Timothy},
  year = {2024},
  url = {https://arxiv.org/abs/2402.02207},
  note = {arXiv:2402.02207}
}

@misc{kumar-etal-2025-polyguard,
  title = {{PolyGuard}: A Multilingual Safety Moderation Tool for 17 Languages},
  author = {Kumar, Priyanshu and Jain, Devansh and Yerukola, Akhila and Jiang, Liwei and Beniwal, Himanshu and Hartvigsen, Thomas and Sap, Maarten},
  year = {2025},
  url = {https://arxiv.org/abs/2504.04377},
  note = {arXiv:2504.04377}
}

@inproceedings{jindal-etal-2025-sage,
  title = {{SAGE}: A Generic Framework for {LLM} Safety Evaluation},
  author = {Jindal, Madhur and Shrawgi, Hari and Agrawal, Parag and Dandapat, Sandipan},
  booktitle = {Proceedings of the 2025 Conference on Empirical Methods in Natural Language Processing: Industry Track},
  year = {2025},
  pages = {11--33},
  publisher = {Association for Computational Linguistics},
  url = {https://aclanthology.org/2025.emnlp-industry.2/},
  doi = {10.18653/v1/2025.emnlp-industry.2}
}

@article{inan-etal-2023-llamaguard,
  title        = {Llama Guard: {LLM}-based Input-Output Safeguard for Human-{AI} Conversations},
  author       = {Inan, Hakan and Upasani, Kartikeya and Chi, Jianfeng and Rando, Javier and
                  Khabsa, Madian and Metz, Luke and Chen, Ruoshi and Sifer, Michael and
                  Szeve, Carly and Kerkez, Victor and others},
  journal      = {arXiv preprint arXiv:2312.06674},
  year         = {2023}
}

@techreport{metallamateam-2024-llamaguard2,
  title        = {Llama Guard 2},
  author       = {{Meta Llama Team}},
  institution  = {Meta},
  note         = {arXiv preprint arXiv:2403.13031},
  year         = {2024}
}

@article{zeng-etal-2024-shieldgemma,
  title        = {{ShieldGemma}: Generative {AI} Content Moderation Based on Gemma},
  author       = {Zeng, Wenjun and Liu, Yuchi and Mullins, Ryan and Peran, Requires and
                  Fernandez, Joe and Hengst, Hamza and others},
  journal      = {arXiv preprint arXiv:2407.21772},
  year         = {2024}
}

@article{gupta-etal-2024-granite,
  title        = {Granite Guardian},
  author       = {Gupta, Inkit and Deshpande, Shubham and Silva, Guilherme and others},
  journal      = {arXiv preprint arXiv:2412.07724},
  year         = {2024}
}

@article{wang-etal-2023-xsafety,
  title        = {All Languages Matter: On the Multilingual Safety of Large Language Models},
  author       = {Wang, Wenxuan and Tu, Zhaopeng and Chen, Chang and Yuan, Youliang and
                  Huang, Jen-tse and Jiao, Wenxiang and Lyu, Michael R.},
  journal      = {arXiv preprint arXiv:2310.00905},
  year         = {2023}
}

@inproceedings{deng-etal-2024-multijail,
  title        = {Multilingual Jailbreak Challenges in Large Language Models},
  author       = {Deng, Yue and Zhang, Wenxuan and Pan, Sinno Jialin and Bing, Lidong},
  booktitle    = {Proceedings of the Twelfth International Conference on Learning Representations (ICLR)},
  year         = {2024}
}

@article{degibert-etal-2024-rtplx,
  title        = {{RTP-LX}: Can {LLMs} Evaluate Toxicity in Multilingual Scenarios?},
  author       = {de Gibert, Ona and Lent, Heather and Bai, Aitor Soroa and others},
  journal      = {arXiv preprint arXiv:2404.14397},
  year         = {2024}
}

@article{ahmadian-etal-2024-aya,
  title        = {The Multilingual Alignment Prism: Aligning Global and Local Preferences to Reduce Harm},
  author       = {Ahmadian, Arash and Ermis, Beyza and Fadaee, Marzieh and others},
  journal      = {arXiv preprint arXiv:2406.18682},
  year         = {2024}
}

@article{ning-etal-2025-linguasafe,
  title        = {{LinguaSafe}: A Comprehensive Multilingual Safety Benchmark for Large Language Models},
  author       = {Ning, Zhiyuan and Gu, Tianle and Song, Jiaxin and Hong, Shixin and Li, Lingyu and Liu, Huacan and Li, Jie and Wang, Yixu and Meng, Lingyu and Teng, Yan and Wang, Yingchun},
  journal      = {arXiv preprint arXiv:2508.12733},
  year         = {2025}
}

@article{qwenteam-2025-qwen3guard,
  title        = {Qwen3Guard Technical Report},
  author       = {{Qwen Team}},
  journal      = {arXiv preprint arXiv:2510.14276},
  year         = {2025}
}

@article{gong-etal-2023-figstep,
  title        = {{FigStep}: Jailbreaking Large Vision-Language Models via Typographic Visual Prompts},
  author       = {Gong, Yichen and Ran, Delong and Liu, Jinyuan and Wang, Conglei and
                  Cong, Tianshuo and Wang, Anyu and Duan, Sisi and Wang, Xiaoyun},
  journal      = {arXiv preprint arXiv:2311.05608},
  year         = {2023}
}

@article{chi-etal-2024-llamaguard3vision,
  title        = {Llama Guard 3 Vision: Safeguarding Human-{AI} Image Understanding Conversations},
  author       = {Chi, Jianfeng and Karn, Ujjwal and Zhan, Hongyuan and Smith, Eric and
                  Rando, Javier and Zhang, Yiming and Plawiak, Kate and
                  Delpierre Coudert, Zacharie and Upasani, Kartikeya and Pasupuleti, Mahesh},
  journal      = {arXiv preprint arXiv:2411.10414},
  year         = {2024}
}

@techreport{metallamateam-2025-llamaguard4,
  title        = {Llama Guard 4},
  author       = {{Meta Llama Team}},
  institution  = {Meta},
  note         = {Model card: \url{https://www.llama.com/docs/model-cards-and-prompt-formats/llama-guard-4/}},
  year         = {2025}
}

@article{rottger-etal-2025-msts,
  title        = {{MSTS}: A Multimodal Safety Test Suite for Vision-Language Models},
  author       = {R{\"o}ttger, Paul and others},
  journal      = {arXiv preprint arXiv:2501.10057},
  year         = {2025}
}

@article{li-etal-2024-rtvlm,
  title        = {Red Teaming Visual Language Models},
  author       = {Li, Mukai and Li, Lei and Yin, Yuwei and Ahmed, Masood and Liu, Zhenguang and Liu, Qi},
  journal      = {arXiv preprint arXiv:2401.12915},
  year         = {2024}
}

@article{rottger-etal-2023-xstest,
  title        = {{XSTest}: A Test Suite for Identifying Exaggerated Safety Behaviours in Large Language Models},
  author       = {R{\"o}ttger, Paul and Kirk, Hannah Rose and Vidgen, Bertie and Attanasio, Giuseppe and Bianchi, Federico and Hovy, Dirk},
  journal      = {arXiv preprint arXiv:2308.01263},
  year         = {2023}
}

@article{yue-etal-2023-mmmu,
  title        = {{MMMU}: A Massive Multi-discipline Multimodal Understanding and Reasoning Benchmark for Expert {AGI}},
  author       = {Yue, Xiang and Ni, Yuansheng and Zhang, Kai and Zheng, Tianyu and Liu, Ruoqi and Zhang, Ge and Stevens, Samuel and Jiang, Dongfu and Ren, Weiming and Sun, Yuxuan and Wei, Cong and Yu, Botao and Yuan, Ruibin and Sun, Renliang and Yin, Ming and Zheng, Boyuan and Yang, Zhenzhu and Liu, Yibo and Huang, Wenhao and Sun, Huan and Su, Yu and Chen, Wenhu},
  journal      = {arXiv preprint arXiv:2311.16502},
  year         = {2023}
}

@inproceedings{mathew-etal-2021-docvqa,
  title        = {{DocVQA}: A Dataset for {VQA} on Document Images},
  author       = {Mathew, Minesh and Karatzas, Dimosthenis and Jawahar, C. V.},
  booktitle    = {Proceedings of the IEEE/CVF Winter Conference on Applications of Computer Vision},
  year         = {2021}
}

@article{kembhavi-etal-2016-ai2d,
  title        = {A Diagram Is Worth A Dozen Images},
  author       = {Kembhavi, Aniruddha and Salvato, Mike and Kolve, Eric and Seo, Minjoon and Hajishirzi, Hannaneh and Farhadi, Ali},
  journal      = {arXiv preprint arXiv:1603.07396},
  year         = {2016}
}

@article{hoover-etal-2025-dynaguard,
  title        = {{DynaGuard}: A Dynamic Guardian Model With User-Defined Policies},
  author       = {Hoover, Monte and Baherwani, Vatsal and Jain, Neel and Saifullah, Khalid and Vincent, Joseph and Jain, Chirag and Rad, Melissa Kazemi and Bruss, C. Bayan and Panda, Ashwinee and Goldstein, Tom},
  journal      = {arXiv preprint arXiv:2509.02563},
  year         = {2025}
}

@inproceedings{zhang-etal-2025-cosa,
  title        = {Controllable Safety Alignment: Inference-Time Adaptation to Diverse Safety Requirements},
  author       = {Zhang, Jingyu and Elgohary, Ahmed and Magooda, Ahmed and Khashabi, Daniel and Van Durme, Benjamin},
  booktitle    = {Proceedings of the Thirteenth International Conference on Learning Representations},
  year         = {2025}
}

\appendix
\section{Fine-tuning Hyperparameters}
\label{app:sft-hyperparams}

\begin{table}[t]
    \centering
    \small
    \caption{Key hyperparameters used for supervised fine-tuning (SFT) of Gemma 3.}
    \label{tab:gemma3-sft-hyperparams}
    \begin{tabular}{ll}
    \toprule
    \textbf{Hyperparameter} & \textbf{Value} \\
    \midrule
    LoRA rank $r$ & 8 \\
    LoRA alpha & 32 \\
    LoRA dropout & 0.05 \\
    Maximum sequence length & 8000 \\
    Per-device train batch size & 2 \\
    Gradient accumulation steps & 8 \\
    Training epochs & 5 \\
    Learning rate & $1.0 \times 10^{-4}$ \\
    Optimizer & AdamW \\
    Learning-rate scheduler & Cosine \\
    Warmup ratio & 0.1 \\
    Maximum gradient norm & 1.0 \\
    Precision & bfloat16 \\
    \bottomrule
    \end{tabular}
\end{table}

\section{Additional Results}
\label{sec:additional-results}
This appendix expands the capability-axis summary in Table~\ref{tab:capability-dashboard}. We first provide the detailed multimodal and text-safety matrix, then the multilingual aggregate and per-language breakdowns, followed by the custom-policy, benign false-positive, and latency details used in the main analysis.

\subsection{Multimodal and Text Safety}

Table~\ref{tab:appendix-core-results} gives the benchmark-level harmful-F1 values behind the multimodal and text-safety rows of Table~\ref{tab:capability-dashboard}. The multimodal rows include only baselines that support image-conditioned moderation; text-only systems are marked unsupported for those rows.

\begin{table*}[t]
\centering
{\setlength{\tabcolsep}{2.3pt}
\begin{tabular}{L{2.9cm}L{3.35cm}C{1.7cm}C{1.1cm}C{1.15cm}C{1.1cm}C{1.55cm}C{1.5cm}}
\hline
Benchmark & Split & \modeltable{} & LG4 & LG3-V & Omni & Qwen3G & Gemma 3-4B \\
\hline
VLGuard & Prompt harmful-F1 & \textbf{0.90} & 0.65 & 0.35 & 0.22 & N/A & 0.75 \\
MM-SafetyBench & Prompt harmful-F1 & \textbf{0.71} & 0.61 & 0.48 & 0.22 & N/A & 0.60 \\
Aegis 2.0 & Prompt harmful-F1 & 0.86 & 0.72 & 0.74 & 0.74 & \textbf{0.87} & 0.81 \\
Aegis 2.0 & Response harmful-F1 & 0.85 & 0.64 & 0.60 & 0.64 & \textbf{0.86} & 0.77 \\
XSTest & Prompt harmful-F1 & 0.85 & 0.83 & 0.85 & 0.73 & \textbf{0.90} & 0.80 \\
XSTest & Response harmful-F1 & 0.87 & 0.88 & 0.85 & 0.65 & \textbf{0.92} & 0.75 \\
WildGuard & Prompt harmful-F1 & 0.85 & 0.74 & 0.76 & 0.57 & \textbf{0.90} & 0.81 \\
WildGuard & Response harmful-F1 & 0.77 & 0.67 & 0.60 & 0.49 & \textbf{0.79} & 0.67 \\
\hline
\end{tabular}}
\caption{Expanded multimodal and text safety harmful-F1 results. N/A indicates that the baseline does not support the corresponding multimodal setting.}
\label{tab:appendix-core-results}
\end{table*}

\subsection{Multilingual Results}

Table~\ref{tab:appendix-multilingual-summary} reports aggregate multilingual scores across benchmarks, while Tables~\ref{tab:appendix-polyguard-language} and~\ref{tab:appendix-linguasafe-language} show the language-level slices used to inspect whether aggregate scores hide uneven coverage.

\begin{table*}[t]
\centering
{\setlength{\tabcolsep}{2.3pt}
\begin{tabular}{L{2.95cm}L{3.65cm}C{1.7cm}C{1.55cm}C{1.45cm}C{1.4cm}C{1.5cm}}
\hline
Benchmark & Metric & \modeltable{} & Qwen3G & PolyG-Q & GG 3.3 & Gemma 3-4B \\
\hline
PolyGuard & Prompt harmful-F1 & 0.80 & \textbf{0.86} & 0.85 & 0.75 & 0.77 \\
PolyGuard & Response harmful-F1 & 0.75 & \textbf{0.78} & 0.75 & 0.65 & 0.68 \\
RTP-LX & Prompt harmful-F1 & 0.89 & \textbf{0.90} & 0.87 & 0.89 & 0.88 \\
MultiJail & Prompt harmful-F1 & 0.95 & \textbf{0.96} & 0.94 & 0.91 & 0.94 \\
XSafety & Prompt harmful-F1 & 0.72 & 0.65 & 0.65 & \textbf{0.74} & 0.68 \\
XSafety & Harmful recall & \textbf{0.56} & 0.51 & 0.50 & 0.38 & 0.51 \\
Aya Red Teaming & Prompt harmful-F1 & 0.97 & \textbf{0.98} & 0.95 & 0.96 & 0.95 \\
Aya Red Teaming & Harmful recall & 0.94 & \textbf{0.98} & 0.95 & 0.92 & 0.91 \\
Aegis Generic & Prompt harmful-F1 & 0.82 & \textbf{0.85} & \textbf{0.85} & 0.76 & 0.77 \\
Aegis Generic & Response harmful-F1 & \textbf{0.84} & 0.83 & 0.80 & 0.76 & 0.78 \\
Aegis Adapted & Prompt harmful-F1 & \textbf{0.97} & 0.86 & 0.96 & 0.89 & 0.90 \\
Aegis Adapted & Response harmful-F1 & \textbf{0.95} & 0.84 & 0.87 & 0.82 & 0.87 \\
LinguaSafe & Average-F1 & 0.71 & \textbf{0.76} & 0.72 & 0.64 & 0.68 \\
\hline
\end{tabular}}
\caption{Expanded multilingual summary results. GG 3.3 denotes Granite Guardian 3.3.}
\label{tab:appendix-multilingual-summary}
\end{table*}

\begin{table*}[t]
\centering
{\setlength{\tabcolsep}{4pt}
\begin{tabular}{L{2.4cm}C{3.0cm}C{3.0cm}C{3.0cm}C{3.0cm}}
\hline
Language & \begin{tabular}[c]{@{}c@{}}\modeltable{}\\prompt\end{tabular} & \begin{tabular}[c]{@{}c@{}}Gemma 3-4B\\prompt\end{tabular} & \begin{tabular}[c]{@{}c@{}}\modeltable{}\\response\end{tabular} & \begin{tabular}[c]{@{}c@{}}Gemma 3-4B\\response\end{tabular} \\
\hline
English & \textbf{0.846} & 0.809 & \textbf{0.772} & 0.669 \\
Arabic & \textbf{0.801} & 0.777 & \textbf{0.767} & 0.655 \\
German & \textbf{0.799} & 0.769 & \textbf{0.762} & 0.679 \\
Spanish & \textbf{0.812} & 0.783 & \textbf{0.753} & 0.679 \\
French & \textbf{0.805} & 0.767 & \textbf{0.747} & 0.679 \\
Hindi & \textbf{0.807} & 0.766 & \textbf{0.742} & 0.696 \\
Japanese & \textbf{0.785} & 0.761 & \textbf{0.728} & 0.664 \\
Thai & \textbf{0.781} & 0.728 & \textbf{0.744} & 0.662 \\
Chinese & \textbf{0.811} & 0.788 & \textbf{0.751} & 0.685 \\
Italian & \textbf{0.804} & 0.762 & \textbf{0.749} & 0.679 \\
Dutch & \textbf{0.800} & 0.743 & \textbf{0.756} & 0.671 \\
Korean & \textbf{0.789} & 0.756 & \textbf{0.747} & 0.699 \\
\hline
\end{tabular}}
\caption{PolyGuard per-language harmful-F1 comparison against the Gemma 3-4B baseline.}
\label{tab:appendix-polyguard-language}
\end{table*}

\begin{table*}[t]
\centering
{\setlength{\tabcolsep}{8pt}
\begin{tabular}{L{2.6cm}C{2.05cm}C{2.05cm}C{2.05cm}C{2.3cm}}
\hline
Language & \modeltable{} & Qwen3G & GG 3.3 & Gemma 3-4B \\
\hline
English & 0.713 & \textbf{0.759} & 0.713 & 0.704 \\
Arabic & 0.691 & \textbf{0.732} & 0.643 & 0.660 \\
Korean & 0.689 & \textbf{0.745} & 0.606 & 0.674 \\
Vietnamese & 0.719 & \textbf{0.779} & 0.591 & 0.676 \\
Chinese & 0.741 & \textbf{0.779} & 0.713 & 0.702 \\
Thai & 0.710 & \textbf{0.742} & 0.526 & 0.662 \\
Russian & 0.722 & \textbf{0.766} & 0.674 & 0.689 \\
\hline
\end{tabular}}
\caption{LinguaSafe per-language Average-F1. GG 3.3 denotes Granite Guardian 3.3. Vietnamese and Russian are outside the 12 explicitly trained languages for \modeltable{}.}
\label{tab:appendix-linguasafe-language}
\end{table*}

\subsection{Custom-Policy Results}

Tables~\ref{tab:dynaguardrail} and~\ref{tab:cosa} expand the custom-policy row of Table~\ref{tab:capability-dashboard}. DynaGuardrail groups results by policy domain, while CoSA tests role and scenario-specific policies; both are reported with direct classification and reasoning on.

\begin{table*}[t]
\centering
{\setlength{\tabcolsep}{8pt}
\begin{tabular}{L{3.8cm}C{1.8cm}C{1.8cm}C{1.8cm}C{2.0cm}}
\hline
Model / setting & Safety & Finance & Tax & Injection \\
\hline
\modeltable{} & 0.91 & 0.84 & 0.86 & \textbf{0.90} \\
\modeltable{}\reasonmark{} & 0.86 & \textbf{0.85} & \textbf{0.89} & 0.88 \\
Granite Guardian 3.3 & 0.91 & 0.83 & 0.83 & 0.87 \\
\hline
\end{tabular}}
\caption{DynaGuardrail custom-policy F1 by policy domain. \reasonmark{} = reasoning on.}
\label{tab:dynaguardrail}
\end{table*}

\begin{table*}[t]
\centering
{\setlength{\tabcolsep}{5pt}
\begin{tabular}{L{3.8cm}C{1.55cm}C{1.75cm}C{1.75cm}C{1.6cm}C{1.6cm}}
\hline
Model / setting & Game & Prosec. & Publish. & Lang. & Film \\
\hline
\modeltable{} & 0.72 & 0.73 & 0.81 & 1.00 & \textbf{0.86} \\
\modeltable{}\reasonmark{} & \textbf{0.83} & \textbf{0.76} & \textbf{0.82} & 1.00 & 0.73 \\
Granite Guardian 3.3 & 0.68 & 0.69 & 0.80 & 1.00 & 0.81 \\
\hline
\end{tabular}}
\caption{CoSA custom-policy F1 by scenario. \reasonmark{} = reasoning on. Prosec., Publish., and Lang. abbreviate the Public Prosecutor, Book Publisher Arab, and Language Learning scenarios.}
\label{tab:cosa}
\end{table*}

\subsection{Benign False Positives and Latency}

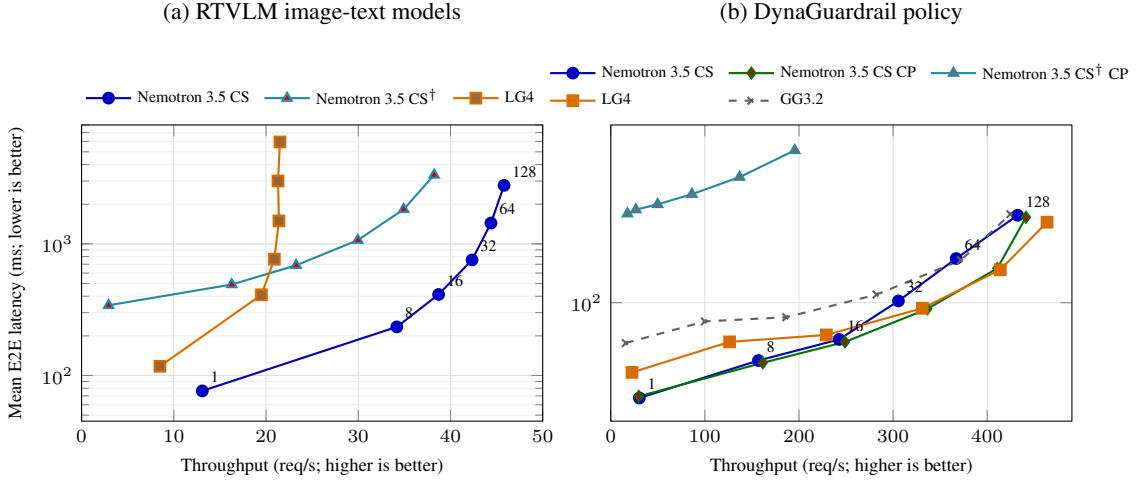
\begin{figure*}[t]
\centering
\small
\begin{minipage}{0.92\textwidth}
\small
\textbf{Reading the curves.} The good direction is down and right: lower latency means faster moderation, while higher throughput means more requests completed per second. Higher concurrency can raise both throughput and latency because the GPU is busier but each request may wait behind more in-flight work.
\end{minipage}
\vspace{0.6em}

\begin{tikzpicture}
\begin{axis}[
    name=rtvlm,
    width=0.48\textwidth,
    height=5.5cm,
    title={(a) RTVLM image-text models},
    xlabel={Throughput (req/s; higher is better)},
    ylabel={Mean E2E latency (ms; lower is better)},
    ymode=log,
    ymin=45,
    ymax=8000,
    xmin=0,
    xmax=50,
    grid=both,
    minor grid style={gray!12},
    major grid style={gray!25},
    legend style={font=\tiny, at={(0.5,1.02)}, anchor=south, draw=none, fill=white, fill opacity=0.95, text opacity=1, legend columns=3, /tikz/every even column/.append style={column sep=0.35em}},
    legend cell align={left},
    tick label style={font=\scriptsize},
    label style={font=\scriptsize},
    title style={font=\small, yshift=3.2em}
]
\addplot+[blue!70!black, mark=*, thick] coordinates {
    (13.09,76.39) (34.18,234.00) (38.71,412.82) (42.34,754.78) (44.39,1438.37) (45.81,2781.24)
};
\addlegendentry{\modeltable{}}
\addplot+[cyan!60!black, mark=triangle*, thick] coordinates {
    (2.93,341.39) (16.28,490.99) (23.26,687.12) (29.95,1066.61) (34.92,1828.68) (38.23,3337.49)
};
\addlegendentry{\modeltable{}\reasonmark{}}
\addplot+[orange!85!black, mark=square*, thick] coordinates {
    (8.50,117.51) (19.50,410.23) (20.90,765.07) (21.40,1494.36) (21.30,3004.16) (21.51,5941.99)
};
\addlegendentry{LG4}
\node[font=\tiny, anchor=south west] at (axis cs:13.09,76.39) {1};
\node[font=\tiny, anchor=south west] at (axis cs:34.18,234.00) {8};
\node[font=\tiny, anchor=south west] at (axis cs:38.71,412.82) {16};
\node[font=\tiny, anchor=south west] at (axis cs:42.34,754.78) {32};
\node[font=\tiny, anchor=south west] at (axis cs:44.39,1438.37) {64};
\node[font=\tiny, anchor=south west] at (axis cs:45.81,2781.24) {128};
\end{axis}

\begin{axis}[
    at={(rtvlm.east)},
    xshift=0.9cm,
    anchor=west,
    width=0.48\textwidth,
    height=5.5cm,
    title={(b) DynaGuardrail policy},
    xlabel={Throughput (req/s; higher is better)},
    ylabel={},
    ymode=log,
    ymin=25,
    ymax=800,
    xmin=0,
    xmax=490,
    grid=both,
    minor grid style={gray!12},
    major grid style={gray!25},
    legend style={font=\tiny, at={(0.5,1.02)}, anchor=south, draw=none, fill=white, fill opacity=0.95, text opacity=1, legend columns=3, /tikz/every even column/.append style={column sep=0.35em}},
    legend cell align={left},
    tick label style={font=\scriptsize},
    label style={font=\scriptsize},
    title style={font=\small, yshift=3.2em}
]
\addplot+[blue!70!black, mark=*, thick] coordinates {
    (30.36,32.79) (156.85,50.81) (242.64,65.05) (305.62,102.04) (367.05,167.65) (432.27,278.15)
};
\addlegendentry{\modeltable{}}
\addplot+[green!45!black, mark=diamond*, thick] coordinates {
    (29.78,33.43) (161.61,49.50) (248.88,63.16) (336.38,92.98) (410.68,149.37) (441.21,271.74)
};
\addlegendentry{\modeltable{} CP}
\addplot+[cyan!60!black, mark=triangle*, thick] coordinates {
    (17.74,283.12) (26.62,297.03) (49.82,315.99) (86.21,355.80) (136.93,434.67) (195.52,594.52)
};
\addlegendentry{\modeltable{}\reasonmark{} CP}
\addplot+[orange!85!black, mark=square*, thick] coordinates {
    (22.57,44.23) (126.25,63.22) (229.11,68.57) (331.09,93.70) (413.86,146.93) (463.59,256.41)
};
\addlegendentry{LG4}
\addplot+[gray!75!black, mark=x, dashed, thick] coordinates {
    (15.98,62.37) (99.96,80.34) (186.49,84.36) (282.40,109.95) (371.91,164.45) (424.60,281.63)
};
\addlegendentry{GG3.2}
\node[font=\tiny, anchor=south west] at (axis cs:30.36,32.79) {1};
\node[font=\tiny, anchor=south west] at (axis cs:156.85,50.81) {8};
\node[font=\tiny, anchor=south west] at (axis cs:242.64,65.05) {16};
\node[font=\tiny, anchor=south west] at (axis cs:305.62,102.04) {32};
\node[font=\tiny, anchor=south west] at (axis cs:367.05,167.65) {64};
\node[font=\tiny, anchor=south west] at (axis cs:432.27,278.15) {128};
\end{axis}
\end{tikzpicture}
\caption{Latency-throughput curves across max-concurrency settings. Each curve connects settings 1, 8, 16, 32, 64, and 128; point labels on the \modeltable{} direct line show those settings. The RTVLM panel includes only image-text-capable models evaluated on image-text inputs. DynaGuardrail points are averages over safety, finance, tax, and injection domains; CP = custom policy and \reasonmark{} = reasoning on.}
\label{fig:latency-throughput}
\end{figure*}

Table~\ref{tab:benign-fpr} and Figure~\ref{fig:latency-throughput} support the deployment-facing rows of Table~\ref{tab:capability-dashboard}: false positives on benign multimodal inputs and serving latency under image-text and custom-policy settings.
Latency benchmarks were run with \texttt{vllm bench serve}\footnote{\url{https://docs.vllm.ai/en/stable/cli/bench/serve/}} on a single NVIDIA H100 GPU.

\begin{table*}[t]
\centering
{\setlength{\tabcolsep}{18pt}
\begin{tabular}{L{4.2cm}C{2.2cm}C{2.2cm}C{2.2cm}}
\hline
Model & MMMU & DocVQA & AI2D \\
\hline
\modeltable{} & 0.030 & 0.060 & 0.001 \\
Llama Guard 4 & 0.061 & 0.015 & 0.001 \\
\hline
\end{tabular}}
\caption{False-positive rate on benign multimodal benchmarks.}
\label{tab:benign-fpr}
\end{table*}

\section{Model Selection}
\label{app:model-selection}
To accomplish the stated goals in the Methodology section, we short-listed the following open-weights model families available in late 2025: (1)~Google Gemma-3 models; (2)~NVIDIA Nemotron VL models; (3)~Meta Llama models; and (4)~Qwen VL models.

Gemma 3-4B satisfies all four requirements simultaneously: it incorporates a native vision encoder enabling joint text-image reasoning; its pretraining data spans over 140 languages, providing strong multilingual grounding; it is released under the Gemma Terms of Use permitting both research and commercial deployment; and at 4 billion parameters it fits within the inference budget of latency-sensitive and edge-deployable services. Among the alternatives we considered, the Nemotron VL and Llama multimodal series both start at at least 9B parameters, which exceeds our target footprint, while Qwen VL models at comparable sizes showed weaker multilingual performance in preliminary experiments.

\section{Human Annotation Setups}
\label{app:annotation-setups}

We employed three annotation project setups that differ in how much material is pre-supplied to annotators. In all three setups the accepted prompt-image pair is sent to a backend of three VLMs to generate a candidate assistant response, and the resulting (prompt, image, response) triple is then labeled for safety by in-house annotators.

\paragraph{Setup 1: Synthetically generated prompts and image captions.}
An SDG pipeline generates a list of prompts paired with image captions describing a relevant visual scene. The annotator searches for a matching image from a list of pre-approved resources. If a suitable image is found it is paired with the generated prompt; if no match is found the annotator may either skip the sample or substitute a contextually appropriate image.

\paragraph{Setup 2: Provided real images only.}
Annotators are given a curated set of images without accompanying prompts. They review each image for relevance to a target safety category, discard images that do not lend themselves to a meaningful safety-relevant prompt, and craft a custom prompt to accompany each accepted image.

\paragraph{Setup 3: No prompt or image data.}
Annotators are given only a list of pre-approved image resources. They are responsible for sourcing an image from those resources, selecting a target safety category, and crafting a prompt that contextualizes the image within that category. This setup places the greatest creative burden on the annotator and is used to collect data for categories where pre-generated seeds are unavailable or low quality.

Intuitively, the first setup seems the most efficient since the annotator is only responsible for finding a relevant image that matches the caption. The third setup provides the most flexibility since the annotator is responsible for sourcing an image and crafting a prompt. The second setup lies somewhere between the first and third setups in terms of efficiency and flexibility. In practice, however we found that the annotation velocity was 25-30\% slower for the first setup compared to the second setup. The annotators found it difficult to find a relevant image that matches the caption and often would have to resort to using a different image.

In all three setups, the labeling of the prompts, images and responses were done by the same annotator. We used a single annotator setup followed by a QA check to ensure the quality of the annotations. Any annotations flagged by QA were sent back to the original annotator for correction. Finally, an audit process randomly sampled 10\% of the annotations as a final check to ensure the quality.

\section{Chat Template and Default Prompt}
\label{app:chat-template}

Listing~\ref{lst:chat-template} shows the chat template used to assemble moderation requests. The template accepts either a free-form custom policy or a custom taxonomy, but not both; otherwise it falls back to the default safety taxonomy. It also appends the category-output control token and the reasoning control token used to toggle direct classification and reasoning-on outputs.

\begin{lstlisting}[style=promptlisting,caption={Chat template and default safety prompt.},label={lst:chat-template}]
{{ bos_token }}
{%- if custom_policy is defined and custom_policy and custom_taxonomy is defined and custom_taxonomy -%}
    {{ raise_exception("Set either custom_policy (Mode A: custom policy) or custom_taxonomy (Mode B: category list only), not both.") }}
{%- endif -%}

{%- if enable_thinking is defined -%}
    {%- set thinking_token = "/think" if enable_thinking else "/no_think" -%}
{%- else -%}
    {%- set thinking_token = "/no_think" -%}
{%- endif -%}

{%- set default_output_format = """User Safety: safe or unsafe rating of the user input
Response Safety: safe or unsafe rating of the assistant response. Omit if no assistant response present.
Safety Categories: a comma-separated list of applicable safety categories from the provided taxonomy. Omit if all safe or not needed.""" -%}

{%- if custom_policy is defined and custom_policy -%}
    {%- set prefix = custom_policy ~ """
<BEGIN CONVERSATION>
user: """ -%}
    {%- set suffix = """<END CONVERSATION>
Output your answer in the following format:

User Safety: safe or unsafe rating of the user input
Response Safety: safe or unsafe rating of the assistant response. Omit if no assistant response present.
Safety Categories: a comma-separated list of applicable safety categories from the custom policy. Omit if all safe or not needed.

Only output the answer in the format specified above after thinking about the user input and assistant response.

Do not include anything other than the output in the above format in your response. """ ~ (request_categories or "/no_categories") ~ " " ~ thinking_token ~ "\nOutput:" -%}
{%- elif custom_taxonomy is defined and custom_taxonomy -%}
    {%- set prefix = """You are a safety classifier for analyzing interactions between humans and LLMs. Given a request from a human user and a response from an LLM assistant, determine:
- Whether the user input is safe or unsafe.
- Whether the LLM assistant response is safe or unsafe.
- Whether the LLM assistant response is a refusal.
- If either the user input or assistant response is unsafe, determine the harmful categories that the input or response falls into.

Taxonomy: The following are the list of safety categories that the input and response may fall into.
""" ~ custom_taxonomy ~ """
<BEGIN CONVERSATION>
user: """
    -%}
    {%- set suffix = """<END CONVERSATION>
Output your answer in the following format:

""" ~ default_output_format ~ """

Only output the answer in the format specified above after thinking about the user input and assistant response.

Do not include anything other than the output in the above format in your response. """ ~ (request_categories or "/no_categories") ~ " " ~ thinking_token ~ "\nOutput:"
    -%}
{%- else -%}
    {%- set prefix = """You are a safety classifier for analyzing interactions between humans and LLMs. Given a request from a human user and a response from an LLM assistant, determine:
- Whether the user input is safe or unsafe.
- Whether the LLM assistant response is safe or unsafe.
- Whether the LLM assistant response is a refusal.
- If either the user input or assistant response is unsafe, determine the harmful categories that the input or response falls into.

Taxonomy: The following are the list of safety categories that the input and response may fall into.
S1: Violence.
S2: Sexual.
S3: Criminal Planning/Confessions.
S4: Guns and Illegal Weapons.
S5: Controlled/Regulated Substances.
S6: Suicide and Self Harm.
S7: Sexual (minor).
S8: Hate/Identity Hate.
S9: PII/Privacy.
S10: Harassment.
S11: Threat.
S12: Profanity.
S13: Needs Caution.
S14: Other.
S15: Manipulation.
S16: Fraud/Deception.
S17: Malware.
S18: High Risk Gov Decision Making.
S19: Political/Misinformation/Conspiracy.
S20: Copyright/Trademark/Plagiarism.
S21: Unauthorized Advice.
S22: Illegal Activity.
S23: Immoral/Unethical.
S24: Economic Harm.
<BEGIN CONVERSATION>
user: """
    -%}
    {%- set suffix = """<END CONVERSATION>
Output your answer in the following format:

""" ~ default_output_format ~ """

Only output the answer in the format specified above after thinking about the user input and assistant response.

Do not include anything other than the output in the above format in your response. """ ~ (request_categories or "/no_categories") ~ " " ~ thinking_token ~ "\nOutput:"
    -%}
{%- endif -%}

{%- set ns = namespace(prompt='', image_token='') -%}

{%- if messages[0]['role'] == 'system' -%}
    {%- if messages[0]['content'] is string -%}
        {%- set first_user_prefix = messages[0]['content'] + '\n\n' -%}
    {%- else -%}
        {%- set first_user_prefix = messages[0]['content'][0]['text'] + '\n\n' -%}
    {%- endif -%}
    {%- set loop_messages = messages[1:] -%}
{%- else -%}
    {%- set first_user_prefix = "" -%}
    {%- set loop_messages = messages -%}

{%- endif -%}

{{ "<start_of_turn>user\n" }}
{%- for message in loop_messages -%}
    {%- if (message['role'] == 'user') != (loop.index0 % 2 == 0) -%}
        {{ raise_exception("Conversation roles must alternate user/assistant/user/assistant/...") }}
    {%- endif -%}
    {%- if (message['role'] == 'assistant') -%}
        {%- set role = "model" -%}
    {%- else -%}
        {%- set role = message['role'] -%}
    {%- endif -%}

    {%- if message['content'] is string -%}
        {%- if (message['role'] == 'user') -%}
            {{ ns.image_token + prefix + (message['content'] | trim) }}
        {%- else -%}
            {{ 'response: agent:  ### Answer: ' + (message['content'] | trim) }}
        {%- endif -%}
    {%- elif message['content'] is iterable -%}
        {%- for item in message['content'] -%}
            {%- if item['type'] == 'image' -%}
                {%- if loop.last -%}
                    {{ '<start_of_image>' + ns.prompt | trim }}
                {%- else -%}
                    {%- set ns.image_token="<start_of_image>" -%}
                {%- endif -%}
            {%- elif item['type'] == 'text' -%}
                {%- if (message['role'] == 'user') -%}
                    {%- if loop.last -%}
                        {{ ns.image_token + prefix + item['text'] | trim }}
                    {%- else -%}
                        {%- set ns.prompt = prefix + item['text'] | trim -%}
                    {%- endif -%}

                {%- else -%}
                    {{ 'response: agent:  ### Answer: ' + item['text'] | trim }}
                {%- endif -%}

            {%- endif -%}
        {%- endfor -%}
    {%- else -%}
        {{ raise_exception("Invalid content type") }}
    {%- endif -%}
    {{ '\n' }}

{%- endfor -%}

{{ suffix + "<end_of_turn>\n<start_of_turn>model\n" }}
\end{lstlisting}

\end{document}